\pdfoutput=1

\documentclass[11pt]{article}

\usepackage[]{emnlp2021}

\usepackage{times}
\usepackage{latexsym}

\usepackage[T1]{fontenc}

\usepackage[utf8]{inputenc}

\usepackage{microtype}

\usepackage{booktabs}
\usepackage{graphicx}
\usepackage{multirow}
\usepackage{outlines}
\usepackage{amsmath}
\usepackage{enumitem}

\title{{E}fficient{BERT}: Progressively Searching Multilayer Perceptron via Warm-up Knowledge Distillation}

\author{Chenhe Dong$^1$, Guangrun Wang$^2$, Hang Xu$^3$, Jiefeng Peng$^4$, \\ 
{\bf Xiaozhe Ren$^3$, Xiaodan Liang$^1$\thanks{\ \ Corresponding author.}} \\
 $^1$ Shenzhen Campus of Sun Yat-sen University $^2$ University of Oxford \\
 $^3$ Huawei Noah’s Ark Lab $^4$ DarkMatter AI Research \\
 {\tt dongchh@mail2.sysu.edu.cn, \{xu.hang,renxiaozhe\}@huawei.com} \\
 {\tt \{wanggrun,jiefengpeng,xdliang328\}@gmail.com}}

\date{}

\begin{document}
\maketitle

\begin{abstract}
Pre-trained language models have shown remarkable results on various NLP tasks. Nevertheless, due to their bulky size and slow inference speed, it is hard to deploy them on edge devices. In this paper, we have a critical insight that improving the feed-forward network (FFN) in BERT has a higher gain than improving the multi-head attention (MHA) since the computational cost of FFN is 2$\sim$3 times larger than MHA. Hence, to compact BERT, we are devoted to designing efficient FFN as opposed to previous works that pay attention to MHA. Since FFN comprises a multilayer perceptron (MLP) that is essential in BERT optimization, we further design a thorough search space towards an advanced MLP and perform a coarse-to-fine mechanism to search for an efficient BERT architecture. Moreover, to accelerate searching and enhance model transferability, we employ a novel warm-up knowledge distillation strategy at each search stage. Extensive experiments show our searched EfficientBERT is 6.9$\times$ smaller and 4.4$\times$ faster than BERT$\rm_{BASE}$, and has competitive performances on GLUE and SQuAD Benchmarks. Concretely, EfficientBERT attains a 77.7 average score on GLUE \emph{test}, 0.7 higher than MobileBERT$\rm_{TINY}$, and achieves an 85.3/74.5 F1 score on SQuAD v1.1/v2.0 \emph{dev}, 3.2/2.7 higher than TinyBERT$_4$ even without data augmentation. The code is released at \url{https://github.com/cheneydon/efficient-bert}.
\end{abstract}

\section{Introduction}

\begin{figure}
\includegraphics[trim=0 0 0 0, width=\columnwidth, clip]{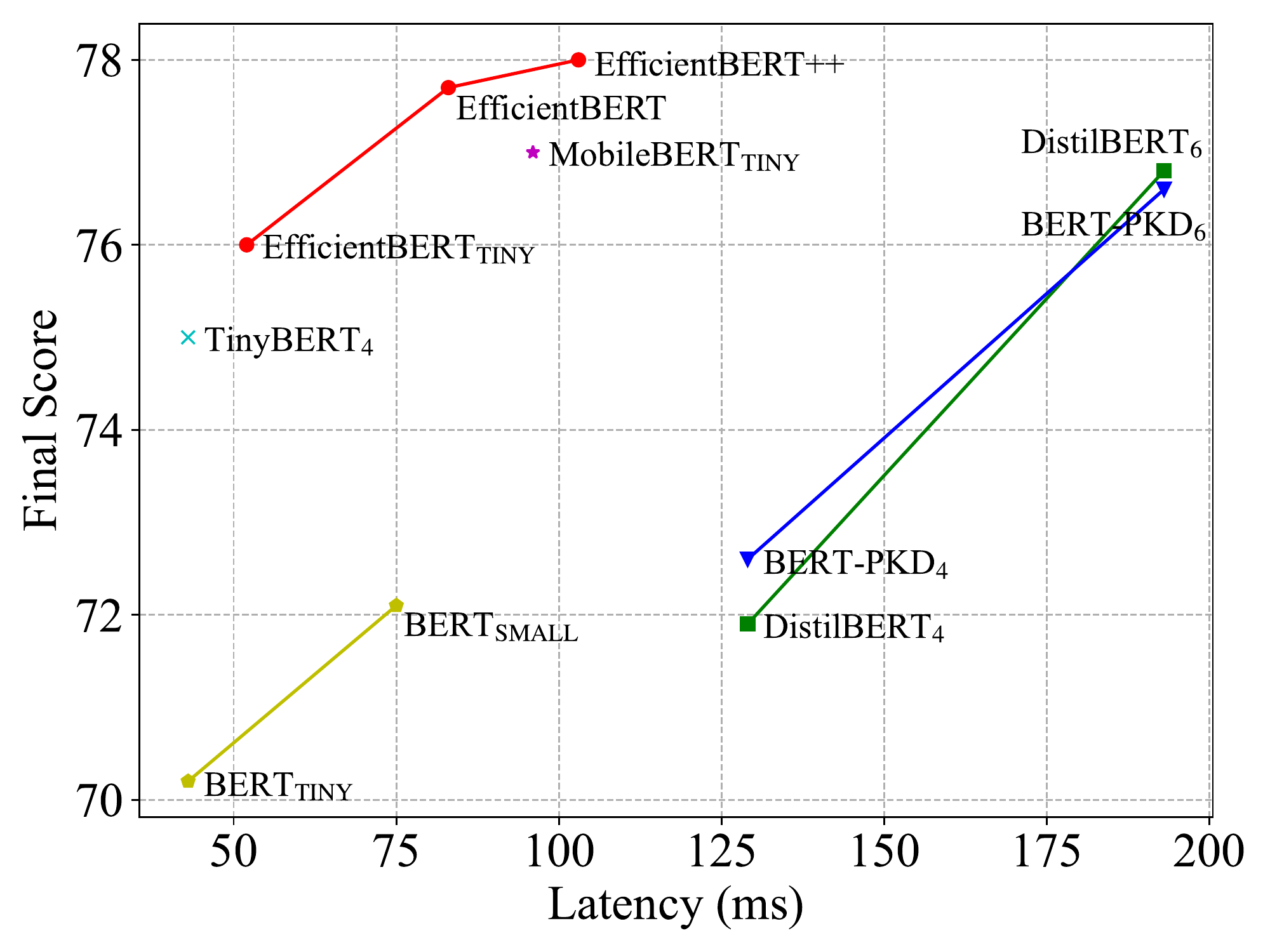}
\caption{Final score vs. latency tradeoff curve. Final score refers to the average score on the GLUE test set.}\label{fig: tradeoff curve}
\end{figure}

Diverse pre-trained language models (PLMs) (e.g., BERT \cite{bert}) have been intensively investigated by designing new pretext tasks, architectures, or attention mechanisms \cite{xlnet, convbert, longformer}. The performances of these PLMs far exceed the traditional methods on a variety of natural language processing (NLP) tasks. Nevertheless, their shortcomings are still evident (including a considerable model size and low inference efficiency), limiting real-world application scenarios.

To alleviate the aforementioned limitations, many model compression methods have been proposed, including quantization, weight pruning, and knowledge distillation (KD) \cite{qbert, are-sixteen-heads-better-than-one, tinybert}. Among them, KD \cite{kd-hinton} that transfers the knowledge from larger teacher models to smaller student models with minimal performance sacrifice is most widely used due to its plug-and-play feasibility and its scalability in the rapid delivery of new models. Specifically, KD allows us to train our own BERT architecture significantly faster than training from scratch. Hence, we adopt KD in this paper. Besides, inspired by the impressive results by neural architecture search (NAS) in vision tasks \cite{mobilenetv3, dna, once-for-all}, adopting NAS to further boost the performance of PLMs or reduce the computational cost has attracted increasing attentions \cite{evolved-transformer, hat, adabert}.

Although considerable progress has been made in the field of KD for PLMs, the compression of the feed-forward network (FFN) has been rarely studied. This contradicts the fact that the computational cost of FFN is 2$\sim$3 times larger than that of the multi-head attention (MHA). In addition, \citet{dong2021attention} have proved that the multilayer perceptron (MLP) in FFN can prevent the undesirable rank collapse caused by self-attention and thus can improve BERT optimization. These motivate us to investigate the nonlinearity of FFN in BERT.

\begin{figure*}
\centering
\includegraphics[trim=0 0 1120 0, width=\textwidth, clip]{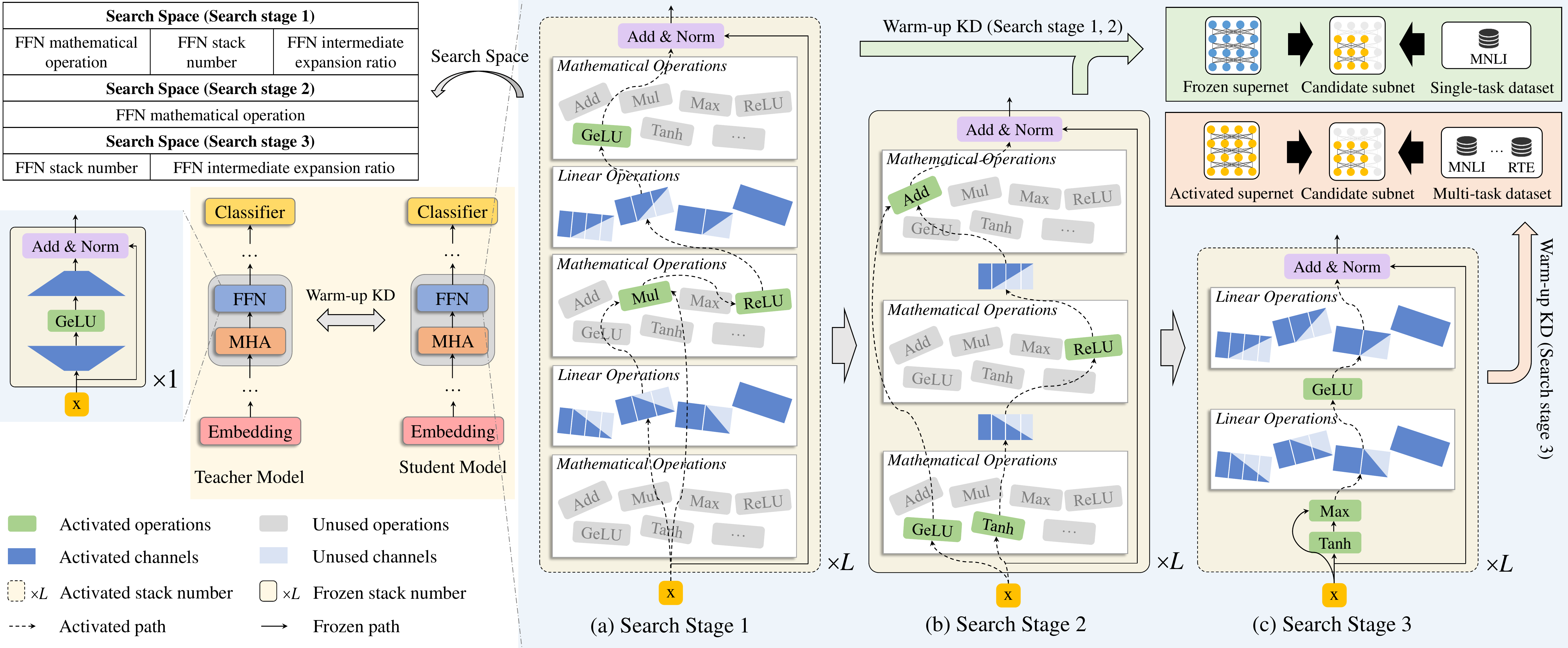}
\caption{An overview of the search procedure of our EfficientBERT. The teacher model is BERT$\rm _{BASE}$ (left), and the search space of our student model is designed towards achieving better nonlinearity of FFN, which contains mathematical operations, stack numbers, and intermediate expansion ratios (right). During searching, we progressively shrink the search space and divide the search process into three stages to conduct NAS ((a)-(c)). Both in the search and retraining stages, we use a novel warm-up knowledge distillation to transfer the teacher model's knowledge (middle). Specifically, each candidate or retrained subnet first inherits the weights from a frozen or activated warmed-up supernet, then conducts pre-training and fine-tuning with single-task or multi-task dataset.}\label{fig: framework}
\end{figure*}

In this paper, we make the first attempt to compress and improve the barely-explored multilayer perceptron (MLP) in FFN and propose a novel coarse-to-fine NAS approach with warm-up KD to find the optimal MLP architectures, aiming to search for a universal small BERT model with competitive performance and strong transferability. Specifically, we design a rich and flexible search space to discover an excellent FFN with maximal nonlinearity and a minimal computational cost. Our search space contains various mathematical operations, stack numbers, and expansion ratios of intermediate hidden size. To efficiently search from our vast search space, we progressively shrink the search space in three stages.
\begin{itemize}[leftmargin=*]
\setlength{\itemsep}{0pt}
\setlength{\parsep}{0pt}
\setlength{\parskip}{0pt}
\item \textbf{Stage 1:} Perform a coarse search to explore the entire search space (i.e., jointly searching the mathematical operations, stack numbers, and expansion ratios)(Figure \ref{fig: framework} (a)). 
\item \textbf{Stage 2:} Fix the stack numbers and expansion ratios, performing a fine-grained search for the optimal mathematical operations (Figure \ref{fig: framework} (b)).
\item \textbf{Stage 3:} Fix the mathematical operations, performing a fine-grained search for optimal stack numbers and expansion ratios (Figure \ref{fig: framework} (c)).
\end{itemize}Even with this elegant coarse-to-fine search strategy, pre-training each candidate model still needs a lot of time to converge during searching. To solve this problem, different from the conventional KD strategy \cite{tinybert}, we propose a warm-up KD strategy to fast transfer the knowledge, where a pre-trained supernet is additionally introduced to perform a joint warm-up for all candidate models. Note that the warm-up strategy in the third stage is slightly different from that of the first two stages. During the first two stages, each candidate model initially inherits its weights from a frozen warmed-up supernet to accelerate searching. But in the third stage, since there is no need to search mathematical operations, an unfrozen warmed-up supernet sharing weights across different candidate models is allowed, i.e., each model can inherit weights from this activated warmed-up supernet for a quick launch and is then trained with weight sharing in a multi-task manner to enhance transferability.

Extensive experimental results show that our searched architecture, named EfficientBERT, is 6.9$\times$ smaller and 4.4$\times$ faster than BERT$\rm _{BASE}$, and has competitive performance. On the test set of GLUE benchmark, EfficientBERT attains an average score of 77.7, which is 0.7 higher than MobileBERT$\rm _{TINY}$, and achieves an F1 score of 85.3/74.5 on the SQuAD v1.1/v2.0 dev dataset, which is 3.2/2.7 higher than TinyBERT$_4$ even without data augmentation.

\section{Related Work}

\paragraph{Compression for Pre-trained Language Models.} For the past few years, pre-trained language models (PLMs) have demonstrated their strong powers on a variety of NLP tasks with the trend of larger and larger model size as well as better results. However, it is hard to deploy them on resource-limited edge devices for practical usage. To solve this problem, many efficient PLMs have been proposed \cite{turc2019wellread, albert}. For example, \citet{turc2019wellread} directly pre-train and fine-tune smaller BERT models. In addition, many compression techniques for PLMs have been proposed recently to reduce the training cost, including quantization, weight pruning, and knowledge distillation (KD) \cite{qbert, poorman-bert, tinybert}. Among them, KD \cite{kd-hinton} is widely used due to its plug-and-play feasibility, which aims to transfer the knowledge from larger teacher models to smaller student models without sacrificing too much performance. For example, BERT-PKD \cite{bert-pkd} jointly distills the intermediate and last layers during fine-tuning. DistilBERT \cite{distilbert} distills the last layers with a triple loss during pre-training. MobileBERT \cite{mobilebert} designs an inverted-bottleneck model structure and progressively transfers the knowledge during pre-training. MiniLM \cite{minilm} performs a deep self-attention distillation during pre-training. TinyBERT \cite{tinybert} introduces a comprehensive Transformer distillation method during pre-training and fine-tuning.

Nevertheless, the compression of the feed-forward network (FFN) has not been well studied, although its computational cost is 2$\sim$3 larger than the multi-head attention (MHA) as pointed out by \citet{squeezebert}. In contrast, compressing FFN is our main focus in this work.

\paragraph{Neural Architecture Search.} Motivated by the success of neural architecture search (NAS) in computer vision \cite{mobilenetv3, dna, once-for-all}, increasing attention has been paid to applying NAS to NLP tasks \cite{evolved-transformer, hat, adabert}, aiming to automatically search for optimal architectures from a vast search space. Evolved Transformer \cite{evolved-transformer} employs NAS to search for a better Transformer architecture with an evolutionary algorithm. HAT \cite{hat} applies NAS to search for efficient hardware-aware Transformer models based on a Transformer supernet. AdaBERT \cite{adabert} searches for task-adaptive small models with KD and differentiable NAS method. NAS-BERT \cite{xu2021nas-bert} proposes a task-agnostic NAS method for adaptive-size model compression, where several acceleration techniques (including block-wise search, search space pruning, and performance approximation) are introduced to speed up the searching process.

Differently, in this paper, we design a comprehensive search space towards the nonlinearity of multilayer perceptron (MLP) in FFN, and propose a novel coarse-to-fine NAS approach with warm-up KD to find the optimal MLP architectures. Unlike AdaBERT, we apply NAS to search for a small universal BERT with competitive performance and strong transferability. And unlike NAS-BERT, we design a much more flexible search space and use warm-up KD with a coarse-to-fine searching paradigm to accelerate searching and enhance model transferability.

\begin{table}
\centering
\small
\caption{Mathematical operations of FFN. Following the original papers of GeLU \cite{Hendrycks2016Bridging_arxiv} and Leaky ReLU \cite{He2015Delving_iccv}, we let $c_1 = 0.5, c_2 = \sqrt{2/\pi}, c_3 = 0.044715, c_4 = 0.01$.}\label{tab: mathematical operations}
\begin{tabular}{lcc}
\toprule
\textbf{Operation}& \textbf{Expression}& \textbf{Arity} \\ 
\midrule
Add& x + y& 2 \\
Mul& x $\times$ y& 2 \\
Max& $\max(x, y)$& 2 \\
GeLU& $c_1x(1 + \mbox{tanh}(c_2(x + c_3 x^3)))$& 1 \\
Sigmoid& $1 / (1 + e^{-x})$& 1 \\
Tanh& $(e^x - e^{-x}) / (e^x + e^{-x})$& 1 \\
ReLU& $\max(x, 0)$& 1 \\
Leaky ReLU& $x \ \mbox{if} \ x \geq 0 \ \mbox{else} \ c_4x$& 1 \\
ELU& $x \ \mbox{if} \ x \geq 0 \ \mbox{else} \ e^x - 1$& 1 \\
Swish& $x / (1 + e^{-x})$& 1 \\
\bottomrule
\end{tabular}
\end{table}

\section{Our EfficientBERT}

We aim at discovering a lightweight MLP architecture with better nonlinearity in each FFN layer, ensuring the searched model can achieve compelling performance. We first present the search space towards better nonlinearity of MLP in FFN, as described in Section \ref{ssec: search space}. Then we propose a novel coarse-to-fine NAS method with warm-up KD as discussed in Section \ref{ssec: coarse-to-fine nas method}.

\subsection{Search Space Design}\label{ssec: search space}

In a standard Transformer layer, there are two main components: a multi-head attention (MHA) and a feed-forward network (FFN). Theoretically, the computation (Mult-Adds) of MHA and FFN is $O(4Ld^2+L^2d)$ and $O(2\times4Ld^2)$ respectively, where $L$ is the sequence length and $d$ is the channel number. As $d$ gets larger, the computation of FFN gets larger than MHA. And as pointed out by previous works \cite{squeezebert}, the latency of MHA and FFN in each layer of BERT$\rm _{BASE}$ accounts for about 30\% and 70\% on a Google Pixel 3 smartphone, and the parameter numbers for MHA and FFN are about 2.4M and 4.7M, respectively. These demonstrate the potential of compressing FFN, i.e., compressing FFN may be more promising than squeezing MHA. In addition, as discussed by \citet{dong2021attention}, the MLP in FFN can prevent an optimization problem, i.e., rank collapse, caused by self-attention; thus, the nonlinearity ability of FFN deserves to be investigated. Hence, our main focus is compression and improvement of FFN.

We then design a search space towards the nonlinearity of MLP in FFN to search for a model with better nonlinearity of FFN and increase the performance. Many factors determine the FFN nonlinearity, such as the mathematical operations and the expansion ratios of intermediate hidden size. Inspired by MobileBERT \cite{mobilebert}, we find that by increasing the stack number of FFN, the model performance can also be remarkably improved. We integrate all of the above factors into our search space, including the mathematical operations, stack numbers, and intermediate expansion ratios of FFN. \emph{(1) Mathematical operation:} We define some primitive operations (including several binary aggregation functions and unary activation functions) and search their different combinations, as shown in Table \ref{tab: mathematical operations}. \emph{(2) Stack number:} The stack number of FFN is selected from \{1, 2, 3, 4\}. \emph{(3) Intermediate expansion ratio:} The intermediate expansion ratio is selected from \{1, 1/2, 1/3, 1/4\}. Note that the stack number and the intermediate expansion ratio are jointly considered to balance the computation cost, e.g., network parameters. We use a directed acyclic graph (DAG) to represent each FFN architecture when searching the mathematical operations. The mathematical operations and linear operations are optionally placed in the intermediate nodes to process the hidden states. More details of our search space can be found in Figure \ref{fig: framework}.

\subsection{Neural Architecture Search}
\label{ssec: coarse-to-fine nas method}

\paragraph{Base Model Design.} As discussed by previous BERT compression works \cite{tinybert, mobilebert}, there are several strategies to reduce the model size, including the embedding factorization and model width/depth reduction. However, most of the recent works only consider part of these strategies. In our work, we design a base model with all these strategies to make a comprehensive compression. Besides, we find that the expansion ratio of intermediate hidden size in FFN contributes a lot to the model size and inference latency. Thus the reduction of the intermediate expansion ratio is also considered. The detailed settings of our base model can be found in Section \ref{ssec: model settings}.

\paragraph{Coarse-to-Fine NAS with Warm-up KD.} To speed up the search in the vast search space, we propose a coarse-to-fine NAS method by progressively shrinking the search space. The search process is divided into three stages where a coarse-grained search is conducted in the first stage to jointly search all of the factors in our search space, and fine-grained searches are conducted in the last two stages to search for partial factors. 

In the first search stage, we jointly search all of the factors in our search space, including the mathematical operations, stack numbers, and intermediate expansion ratios. We use a DAG computation graph described in \S \ref{ssec: search space} to represent each MLP architecture. The initial search candidates are based on our base model, but different stack numbers and intermediate expansion ratios of FFN are allowed.

During searching, each candidate model is first sampled by a learnable sampling decision tree as proposed in LaNAS \cite{lanas}. Then warm-up KD is employed on each candidate model to accelerate the search process. Since we need to search for the mathematical operations, we cannot share the weights of different candidate models to avoid the potential interference problems. Instead, we first build a warmed-up supernet based on our base model with the maximum FFN stack number and intermediate expansion ratio in our search space. The supernet is pre-trained entirely (i.e., complete graph) with KD. The weights of the supernet are then frozen. When training each candidate model, we first inherit its weights from the supernet. Precisely, the weights of each stacked FFN are sliced from bottom to top layer; and the weights of each linear operation are sliced from left to right channel. After that, we only need to pre-train and fine-tune each model for a few steps via KD to adjust the inherited weights. This significantly reduces the search cost.

In the second search stage, to discover more diversified mathematical operations and evaluate their effects, we search them individually with the same method in the first search stage. The initial search candidates are built upon the searched model of the first search stage (i.e., we fix the stack numbers and expansion ratios). The sampling and KD strategies are the same as the first search stage.

In the third search stage, we jointly search the stack numbers and intermediate expansion ratios in the search space to explore their potentials further. The initial search candidates are based on the second search stage's searched model; the searched mathematical operations are fixed, but different stack numbers and intermediate expansion ratios of FFN are allowed. We also apply warm-up KD to accelerate the searching. Specifically, we first warm up the supernet entirely (i.e., complete graph) via KD again but do not freeze its weights. Then we share the weights of different candidate models (i.e., subgraphs of the supernet) during pre-training and fine-tuning to make acceleration. Each candidate model is sampled uniformly. Compared with the first two search stages, the search cost is dramatically reduced, enabling us to leverage more downstream datasets to enhance the model transferability. Inspired by MT-DNN \cite{mtdnn}, each candidate model is fine-tuned in a multi-task manner on different categories of downstream tasks. The weights of the embedding and Transformer layers for all tasks are shared, while those of the prediction layers are different.

\paragraph{Warm-up KD Formulations.} In our warm-up KD, each candidate/retrained model initially inherits the weights from a warmed-up supernet. We use BERT$\rm _{BASE}$ \cite{bert} as the teacher model. Following TinyBERT \cite{tinybert}, we jointly distill the attention matrices, Transformer-layer outputs, embeddings, and predicted logits between the student and teacher models. In detail, the attention loss at the $m$-th student layer $\mathcal{L}_{attn}^m$ is calculated by the mean square error (MSE) loss as:\begin{equation}
\mathcal{L}_{attn}^m = \frac{1}{h} \sum_{i = 1}^h \mbox{MSE}(\mathbf{A}_{i,m}^S, \mathbf{A}_{i,n}^T), 
\end{equation}where $\mathbf{A}_{i,m}^S$ and $\mathbf{A}_{i,n}^T$ refer to the $i$-th head of attention matrices at $m$-th student layer and its matching $n$-th teacher layer, respectively, and $h$ is the number of attention heads. The Transformer-layer output loss at the $m$-th student layer $\mathcal{L}_{hidn}^m$ and the embedding loss $\mathcal{L}_{embd}$ can be formulated as:\begin{equation}
\left\{
\begin{aligned}
\mathcal{L}_{hidn}^m &= \mbox{MSE}(\mathbf{H}_m^S \mathbf{W}_h, \mathbf{H}_n^T)\\
\mathcal{L}_{embd} &= \mbox{MSE}(\mathbf{E}^S \mathbf{W}_e, \mathbf{E}^T)
\end{aligned},
\right.
\end{equation}where $\mathbf{H}_m^S$ and $\mathbf{H}_n^T$ are the Transformer-layer outputs at $m$-th student layer and its matching $n$-th teacher layer, respectively. $\mathbf{E}$ is the embedding, and two learnable transformation matrices $\mathbf{W}_h$ and $\mathbf{W}_e$ are applied to align the mismatch dimensions between the student and teacher models. Moreover, the prediction loss $\mathcal{L}_{pred}$ calculated by the soft cross-entropy (CE) loss can be formulated as:\begin{equation}
\mathcal{L}_{pred} = \mbox{CE}(\mathbf{z}^S / t, \mathbf{z}^T / t),
\end{equation}where $\mathbf{z}$ is the predicted logits vector, and $t$ is the temperature value. Finally, we combine all of the above losses and derive the overall KD loss as:
\begin{equation}
\mathcal{L} = \sum_{m=1}^{M} (\mathcal{L}_{attn}^m + \mathcal{L}_{hidn}^m) + \mathcal{L}_{embd} + \gamma \mathcal{L}_{pred},
\label{eq: overall loss}
\end{equation}where $M$ is the number of Transformer layers in the student model, $\gamma$ controls the weight of the prediction loss $\mathcal{L}_{pred}$.

\begin{table*}
\centering
\caption{Results on the test set of GLUE benchmark. The architectures of different models are as follows. BERT$_{\rm TINY}$ \& TinyBERT$_4$: ($M$=4, $d$=312, $d_i$=1200); BERT$_{\rm SMALL}$: ($M$=4, $d$=512, $d_i$=2048); BERT-PKD$_4$ \& DistilBERT$_4$: ($M$=4, $d$=768, $d_i$=3072); BERT-PKD$_6$ \& DistilBERT$_6$: ($M$=6, $d$=768, $d_i$=3072). The latency is the average inference time over 100 runs on a single GPU with a batch size of 128.}\label{tab: glue results}
\resizebox{\textwidth}{!}{
\begin{tabular}{l|cc|cccccccc|c}
\hline
\textbf{Model}& \textbf{\#Params}& \textbf{Latency}& \textbf{MNLI-m/mm}& \textbf{QQP}& \textbf{QNLI}& \textbf{SST-2}& \textbf{CoLA}& \textbf{STS-B}& \textbf{MRPC}& \textbf{RTE}& \textbf{Avg} \\
\hline
BERT$\rm _{BASE}$ (Google)& 108.9M& 362ms& 84.6/83.4& 71.2& 90.5& 93.5& 52.1& 85.8& 88.9& 66.4& 79.6 \\
BERT$\rm _{BASE}$ (Teacher)& 108.9M& 362ms& 84.8/83.8& 71.6& 91.3& 93.1& 53.9& 85.3& 89.2& 68.9& 80.2 \\
\hline
BERT$\rm _{TINY}$ \cite{turc2019wellread}& 14.5M& 43ms& 75.4/74.9& 66.5& 84.8& 87.6& 19.5& 77.1& 83.2& 62.6& 70.2 \\
BERT$\rm _{SMALL}$ \cite{turc2019wellread}& 28.8M& 75ms& 77.6/77.0& 68.1& 86.4& 89.7& 27.8& 77.0& 83.4& 61.8& 72.1 \\
BERT-PKD$_4$ \cite{bert-pkd}& 52.8M& 129ms& 79.9/79.3& 70.2& 85.1& 89.4& 24.8& 79.8& 82.6& 62.3& 72.6 \\
BERT-PKD$_6$ \cite{bert-pkd}& 67.0M& 193ms& 81.5/81.0& 70.7& 89.0& 92.0& 43.5& 81.6& 85.0& 65.5& 76.6 \\
DistilBERT$_4$ \cite{distilbert}& 52.8M& 129ms& 78.9/78.0& 68.5& 85.2& 91.4& 32.8& 76.1& 82.4& 54.1& 71.9 \\
DistilBERT$_6$ \cite{distilbert}& 67.0M& 193ms& 82.6/81.3& 70.1& 88.9& \textbf{92.5}& \textbf{49.0}& 81.3& 86.9& 58.4& 76.8 \\
TinyBERT$_4$ \cite{tinybert}& 14.5M& 43ms& 81.8/80.7&  69.6&  87.7&  91.2&  27.2&  83.0&  88.5&  64.9& 75.0 \\
MobileBERT$\rm _{TINY}$ \cite{mobilebert}& 15.1M& 96ms& 81.5/81.6& 68.9& 89.5& 91.7& 46.7& 80.1& 87.9& 65.1& 77.0 \\
\hline
$\hbox{EfficientBERT}_{\rm TINY}$& 9.4M& 52ms& 82.4/81.0& 70.3& 88.5& 91.2& 37.5& 80.9& 87.8& 64.6& 76.0 \\
EfficientBERT w/o Warm-up KD& 15.7M& 83ms& 83.1/82.0& 71.0& 89.5& 90.8& 42.1& 82.1& 88.4& 67.2& 77.4 \\
EfficientBERT& 15.7M& 83ms& \textbf{83.3}/82.3& 71.0& 90.2& 92.1& 43.8& 82.9& 88.2& 65.7& 77.7 \\
EfficientBERT+& 15.7M& 83ms& 83.0/82.3& \textbf{71.2}& 89.3& 92.4& 38.1& \textbf{85.1}& \textbf{89.9}& \textbf{69.4}& 77.9 \\
EfficientBERT++& 16.0M& 103ms& 83.0/\textbf{82.5}& \textbf{71.2}& \textbf{90.6}& 92.3& 42.5& 83.6& 88.9& 67.8& \textbf{78.0} \\
\hline
\end{tabular}}
\end{table*}

\begin{table*}
\centering
\caption{Results on the dev set of GLUE benchmark compared with other NAS methods. $\dagger$ indicates the results with data augmentation.}\label{tab: glue dev results}
\resizebox{0.9\textwidth}{!}{
\begin{tabular}{l|c|cccccccc|c}
\hline
\textbf{Model}& \textbf{\#Params}& \textbf{MNLI-m}& \textbf{QQP}& \textbf{QNLI}& \textbf{SST-2}& \textbf{CoLA}& \textbf{STS-B}& \textbf{MRPC}& \textbf{RTE}& \textbf{Avg} \\
\hline
AdaBERT \cite{adabert} $\dagger$& 6.4$\sim$9.5M& 81.3& 70.5& 87.2& \textbf{91.9}& -& -& 84.7& 64.1& - \\
NAS-BERT$_{10}$ \cite{xu2021nas-bert}& 10M& 76.4& 88.5& 86.3& 88.6& 34.0& 84.8& 79.1& 66.6& 75.5 \\
NAS-BERT$_{30}$ \cite{xu2021nas-bert}& 30M& 81.0& \textbf{90.2}& 88.4& 90.5& 48.7& \textbf{87.6}& 84.6& \textbf{71.8}& 80.3 \\ 
\hline
$\hbox{EfficientBERT}_{\rm TINY}$& 9.4M& 81.7& 86.7& 89.3& 90.1& 39.1& 79.9& 90.1& 63.2& 77.5 \\
EfficientBERT& 15.7M& \textbf{83.1}& 87.3& \textbf{90.4}& 91.3& \textbf{50.2}& 82.5& \textbf{91.5}& 66.8& \textbf{80.4} \\
\hline
\end{tabular}}
\end{table*}

\section{Experiment}
This section demonstrates the superior performance and transferability of our EfficientBERT on a wide range of downstream tasks.

\subsection{Datasets}

We evaluate our model on two standard benchmarks for natural language understanding, i.e., the General Language Understanding Evaluation (GLUE) benchmark \cite{glue} and the Stanford Question Answering Dataset (SQuAD). The GLUE benchmark contains nine classification datasets, including MNLI \cite{mnli}, QQP \cite{qqp}, QNLI \cite{squad}, SST-2 \cite{sst-2}, CoLA \cite{cola}, STS-B \cite{sts-b}, MRPC \cite{mrpc}, RTE \cite{rte}, and WNLI \cite{wnli}. The SQuAD task aims to predict the answer text span of the given question in a Wikipedia passage, which contains two datasets: SQuAD v1.1 \cite{squad} and SQuAD v2.0 \cite{squad2.0}. The metrics can be found in \citet{glue} and \citet{squad}.

\subsection{Model Settings}\label{ssec: model settings}

The embedding factorization strategy of our base model is the same as MobileBERT \cite{mobilebert}, the number of Transformer layers $M$ is set to 6, the hidden size of the model $d$ is set to 540, and the intermediate expansion ratio of FFN is set to 1 with intermediate hidden size $d_i$ of 540. The remaining structures are the same as BERT$\rm _{BASE}$.

We retrain our searched model of the third search stage by employing the warm-up KD method used in the first two search stages described in Section \ref{ssec: coarse-to-fine nas method}, referring to as EfficientBERT. EfficientBERT+ is obtained by inheriting the weights of EfficientBERT from the multi-task fine-tuned supernet and then directly fine-tune on each downstream task. Moreover, to verify the importance of model depth, we extend our EfficientBERT from 6 layers to 12 layers by affinely repeating each layer in EfficientBERT twice and shrink the hidden size from 540 to 360, forming EfficientBERT++. The weights are initially inherited from the warmed-up supernet of EfficientBERT in the same manner. In addition, to ensure a fair comparison with TinyBERT$_4$, we further shrink the hidden size of our EfficientBERT from 540 to 360, forming EfficientBERT$_{\rm TINY}$, which has similar latency with TinyBERT$_4$.\footnote{Our searched model, i.e., EfficientBERT, can be seen in Figure \ref{fig: appendix_searched_models} of the Appendix \ref{sec: appendix_1}.}

\begin{table}
\centering
\caption{Results on the SQuAD dev datasets. The architectures of MiniLM$_4$ and MiniLM$_6$ are ($M$=4, $d$=384, $d_i$=1536) and ($M$=6, $d$=384, $d_i$=1536), respectively. $\dagger$ indicates the results with data augmentation.}\label{tab: squad results}
\resizebox{\columnwidth}{!}{
\begin{tabular}{l|c|cc}
\hline
\multirow{2}{*}{\textbf{Model}}& \multirow{2}{*}{\textbf{\#Params}}& \textbf{SQuAD v1.1}& \textbf{SQuAD v2.0} \\
~& ~& \textbf{EM/F1}& \textbf{EM/F1} \\
\hline
BERT$\rm _{BASE}$ (Google)& 108.9M& 80.8/88.5& -/- \\
BERT$\rm _{BASE}$ (Teacher)& 108.9M& 80.5/88.2& 74.8/77.7 \\
\hline
BERT-PKD$_4$ \cite{bert-pkd}& 52.8M& 70.1/79.5& 60.8/64.6 \\
BERT-PKD$_6$ \cite{bert-pkd}& 67.0M& 77.1/85.3& 66.3/69.8 \\
DistilBERT$_4$ \cite{distilbert}& 52.8M& 71.8/81.2& 60.6/64.1 \\
DistilBERT$_6$ \cite{distilbert}& 67.0M& 78.1/86.2& 66.0/69.5 \\
TinyBERT$_4$ \cite{tinybert} $\dagger$& 14.5M& 72.7/82.1& 68.2/71.8 \\
MiniLM$_4$ \cite{minilm}& 19.3M& -/-& -/69.7 \\
MiniLM$_6$ \cite{minilm}& 22.9M& -/-& -/72.7 \\
\hline
$\hbox{EfficientBERT}_{\rm TINY}$& 9.4M& 74.8/83.6& 68.6/71.9 \\
EfficientBERT& 15.7M& 77.0/85.3& 71.4/74.5 \\
EfficientBERT++& 16.0M& \textbf{78.3}/\textbf{86.5}& \textbf{73.0}/\textbf{76.1} \\
\hline
\end{tabular}}
\end{table}

\begin{table*}
\centering
\caption{Results of searched models at different search stages on the GLUE test set. Wiki and Books refer to the pre-training corpora of English Wikipedia and BooksCorpus, respectively.}\label{tab: candidate model comparison}
\resizebox{\textwidth}{!}{
\begin{tabular}{l|c|cccccccc|c}
\hline
\textbf{Model (Pre-train Dataset)}& \textbf{\#Params}& \textbf{MNLI-m/mm}& \textbf{QQP}& \textbf{QNLI}& \textbf{SST-2}& \textbf{CoLA}& \textbf{STS-B}& \textbf{MRPC}& \textbf{RTE}& \textbf{Avg} \\
\hline
Base Model (Wiki)& 15.3M& 82.5/81.6& 71.0& 89.0& 91.4& 37.3& 82.1& 86.1& 65.8& 76.3 \\
Search Stage 1 (Wiki)& 15.4M& 82.8/82.0& 71.0& 89.7& 91.8& 37.4& 82.2& 87.7& 65.3& 76.7 \\
Search Stage 2 (Wiki)& 15.4M& 82.8/82.3& 70.9& 89.8& \textbf{92.2}& 38.3& 82.1& \textbf{88.5}& 65.7& 77.0 \\
\hline
Search Stage 2 (Wiki+Books)& 15.4M& 82.8/82.0& \textbf{71.1}& 89.7& 92.1& 42.5& 82.2& 88.2& \textbf{66.3}& 77.4 \\
EfficientBERT (Wiki+Books)& 15.7M& \textbf{83.3}/\textbf{82.3}& 71.0& \textbf{90.2}& 92.1& \textbf{43.8}& \textbf{82.9}& 88.2& 65.7& \textbf{77.7} \\
\hline
\end{tabular}}
\end{table*}

\begin{table}
\centering
\caption{Effectiveness comparison between single-stage searching and our coarse-to-fine NAS method.}\label{tab: effectiveness comparison}
\resizebox{\columnwidth}{!}{
\begin{tabular}{lccc}
\hline
\textbf{Method}& \textbf{Best Score}& \textbf{\#Searched Arch}& \textbf{Search Cost} \\
\hline
Single stage& 76.7& 2,700& 84 GPU days \\
\hline
Search stage 1& 76.7& 1,900& 54 GPU days \\
Search stage 1, 2& 77.0& 2,000& 56 GPU days \\
Coarse-to-Fine NAS& 77.7& 5,000& 58 GPU days \\
\hline
\end{tabular}}
\end{table}

\begin{table*}
\centering
\caption{Results of our EfficientBERT with different base models on the GLUE test set.}\label{tab: ablation of architecture transfer}
\resizebox{\textwidth}{!}{
\begin{tabular}{l|c|cccccccc|c}
\hline
\textbf{Model (Base Model)}& \textbf{\#Params}& \textbf{MNLI-m/mm}& \textbf{QQP}& \textbf{QNLI}& \textbf{SST-2}& \textbf{CoLA}& \textbf{STS-B}& \textbf{MRPC}& \textbf{RTE}& \textbf{Avg} \\
\hline
TinyBERT$_6$& 67.0M& 83.8/83.2& 71.4& 89.8& 92.0& 38.8& 83.1& 89.0& 65.8& 77.4 \\
EfficientBERT (TinyBERT$_6$)& 70.1M& \textbf{84.1}/\textbf{83.2}& \textbf{71.4}& \textbf{90.4}& \textbf{92.6}& \textbf{46.2}& \textbf{83.7}& \textbf{89.0}& \textbf{67.7}& \textbf{78.7} \\
\hline
\end{tabular}}
\end{table*}

\subsection{Implementation Details}

In the first two search stages, the frozen supernet sliced by each candidate model is pre-trained for ten epochs, and we use 2\% of the English Wikipedia corpus to pre-train each candidate model for one epoch. During fine-tuning, we use the first 10\% training set of MNLI to train each model for three epochs and the last 1\% training set for evaluation. In the third search stage, the activated supernet is pre-trained and fine-tuned for ten epochs, and each candidate model is optimized for one step. We use the entire corpora of English Wikipedia and BooksCorpus as the pre-training data, the combination of 90\% training set of each downstream GLUE task as the fine-tuning data, and the rest 10\% training set of MNLI as the evaluation data. The batch size at each search stage is set to 256. The learning rates for pre-training and fine-tuning at each stage are set to 1e-4 and 4e-4, respectively.

During retraining, each searched model is first pre-trained for ten epochs based on the inherited weights from the warmed-up supernet and is then fine-tuned on downstream tasks for ten epochs except for CoLA. Note that CoLA is fine-tuned for 50 epochs following the widely-used protocol. The batch sizes for pre-training and fine-tuning are set to 256 and 32, respectively. The learning rate for pre-training is set to 1e-4. The learning rates for fine-tuning on GLUE and SQuAD datasets are set to 5e-5 and 1e-4, respectively.

In all of our experiments, $\gamma$ is set to 0 and 1 for pre-training and fine-tuning, respectively. $t$ is set to 1. The maximum sequence length is set to 128. We use Adam with $\beta_1 = 0.9$, $\beta_2 = 0.999$, L2 weight decay of 0.01, warm-up proportion of 0.1, and linear decay of the learning rate.

\subsection{Results on GLUE}

We compare our searched models with BERT$_{\rm TINY}$, BERT$\rm _{SMALL}$ \cite{turc2019wellread} and several state-of-the-art compressed BERT models, including BERT-PKD \cite{bert-pkd}, DistilBERT \cite{distilbert}, TinyBERT$_4$ \cite{tinybert}, and MobileBERT$\rm _{TINY}$ \cite{mobilebert}. For a fair comparison, TinyBERT$_4$ is re-implemented by removing the data augmentation and fine-tuning from the official general distillation weights\footnote{We use the 2nd version from \url{https://github.com/huawei-noah/Pretrained-Language-Model/tree/master/TinyBERT}}. The experimental results on the test set of GLUE benchmark are listed in Table \ref{tab: glue results} and Figure \ref{fig: tradeoff curve}.

From the results in Table \ref{tab: glue results}, we can observe that: (1) Our EfficientBERT is 6.9$\times$ smaller and 4.4$\times$ faster than BERT$\rm _{BASE}$ and has achieved a competitive average GLUE score of 77.7, which is 0.7 higher than its counterpart MobileBERT$\rm _{TINY}$. (2) Our EfficientBERT+ has better transferability than EfficientBERT across different GLUE tasks with an improvement of 0.2 on the average score, demonstrating the effectiveness of our multi-task training strategy in the third search stage. (3) Our EfficientBERT++ has achieved state-of-the-art performance, which outperforms MobileBERT$\rm _{TINY}$ by 1.0 on the average score. (4) Our EfficientBERT$_{\rm TINY}$ outperforms TinyBERT$_4$ by a 1.0 average score with fewer parameters and similar latency. (5) Without our warm-up KD during retraining, i.e., pre-training the model from scratch rather than from the warmed-up supernet, the average score of EfficientBERT decreases by 0.3, demonstrating the advantage of retraining with our warm-up KD. And from the results in Figure \ref{fig: tradeoff curve}, we can see that all of our searched models outperform other compared models with similar or lower latency.

Furthermore, to verify the effectiveness of our proposed NAS method, we compare with several related NAS methods on the GLUE dev set, including AdaBERT \cite{adabert} and NAS-BERT \cite{xu2021nas-bert}. The results are shown in Table \ref{tab: glue dev results}. As can be seen, with similar parameters, our EfficientBERT$_{\rm TINY}$ has better performance than AdaBERT and NAS-BERT$_{10}$; and our EfficientBERT outperforms NAS-BERT$_{30}$ even with much fewer parameters. These results demonstrate the superiority of our NAS method.

\begin{figure}
\includegraphics[trim=10 30 80 70, width=\columnwidth, clip]{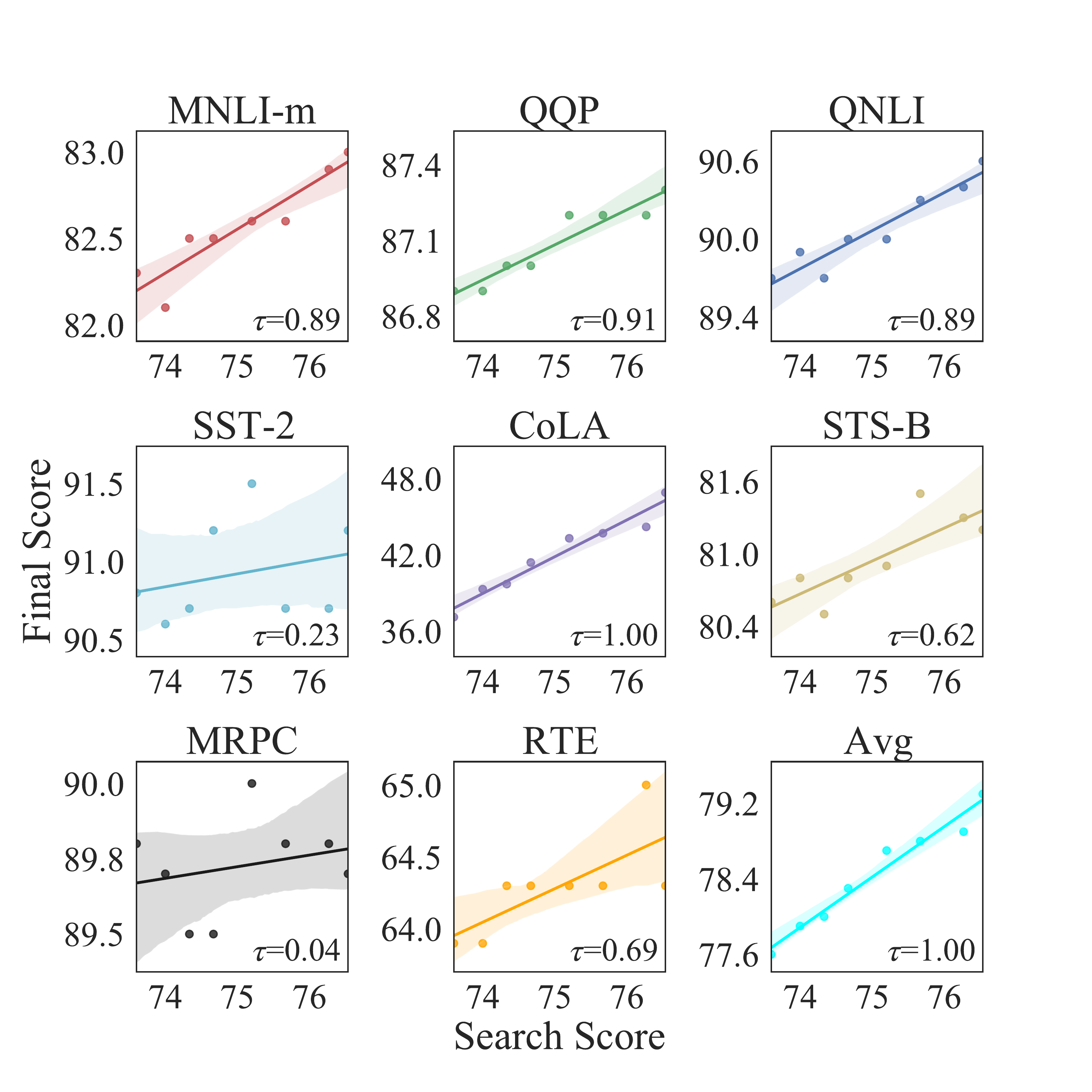}
\caption{Results of model ranking correlation between the search and retraining phases on the GLUE dev set in the first search stage.}\label{fig: model ranking}
\end{figure}

\subsection{Results on SQuAD}

To measure the transferability of our searched models across different types of tasks, we further evaluate our models on SQuAD dev datasets, as shown in Table \ref{tab: squad results}. We choose BERT-PKD, DistilBERT, TinyBERT$_4$, and MiniLM \cite{minilm} as the baseline models. From the results, we can see that our EfficientBERT still achieves competitive performances, which outperforms TinyBERT$_4$ by 3.2/2.7 F1 score on SQuAD v1.1/v2.0 dev dataset even without data augmentation, and surpasses MiniLM$_6$ by 1.8 F1 score on SQuAD v2.0 dev dataset. Besides, our EfficientBERT$_{\rm TINY}$ can also outperform TinyBERT$_4$ on both SQuAD dev datasets. These results indicate the strong performance and transferability of our searched models.

\subsection{Discussion}

\paragraph{Effectiveness of Coarse-to-Fine NAS Method.} To measure the effectiveness of our coarse-to-fine NAS method, we first compare the performances of the searched models at different search stages on the GLUE test set in Table \ref{tab: candidate model comparison}. It can be observed that the searched model in the first search stage has better performance than our base model, which proves the effectiveness of the coarse-grained NAS process. And from the first to the third search stages, the performances of the searched models are gradually enhanced, which shows the effectiveness of the fine-grained strategies and the necessity of each factor in our search space.

Then we compare the effectiveness between single-stage searching and our coarse-to-fine NAS method in Table \ref{tab: effectiveness comparison}. As shown, our coarse-to-fine NAS method has higher efficiency than single-stage searching, saving 26 GPU days. It can also search for 2,300 more architectures and observe better architecture with a higher GLUE test score.

\begin{figure}
\includegraphics[trim=10 10 10 10, width=0.9\columnwidth, clip]{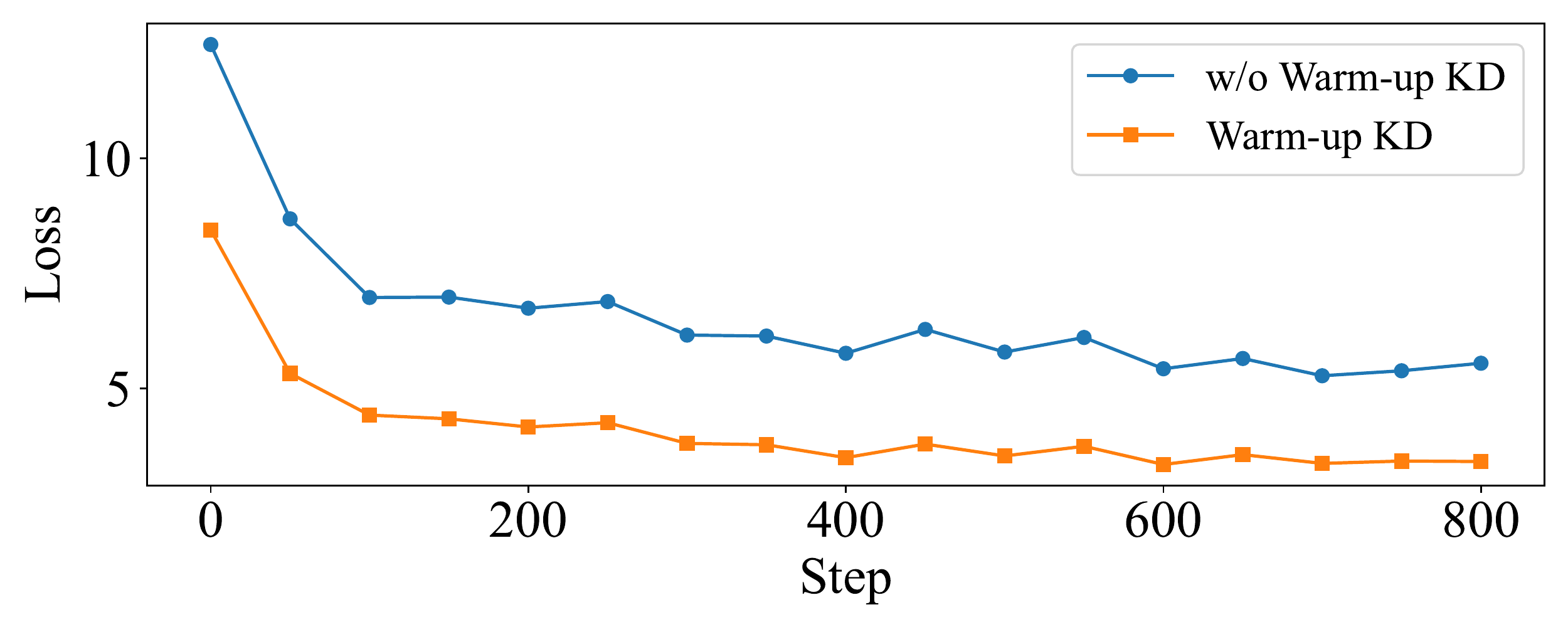}
\caption{Efficiency comparison results between searching with and without our warm-up KD.}\label{fig: warm-up kd effciency}
\end{figure}

\paragraph{Effectiveness of Warm-up KD.} To evaluate the model ranking effectiveness of our warm-up KD method between the search and retraining phases, we first randomly sample eight candidate models in the first search stage, whose search scores range from 77.6 to 79.3. Then we retrain each model and obtain its final score on the GLUE dev set, as shown in Figure \ref{fig: model ranking}. The Kendall Tau $\tau$ \cite{kendall} for each downstream task is also calculated. From the results, we can see that the search and retraining phases have strong positive correlations on most downstream tasks, demonstrating the strong ranking capability of the warm-up KD strategy.

Next, to test the efficiency, we compare the fine-tuning losses of our base model in the first search stage between searching with and without our warm-up KD strategy, as shown in Figure \ref{fig: warm-up kd effciency}. From the results, we can observe that the loss with our warm-up KD can reach a lower value with much fewer steps.

\paragraph{Transferability across Different Base Models.} To test the transferability of our EfficientBERT across different base models, we replace the architecture of TinyBERT$_6$ with that of the EfficientBERT and evaluate it on the GLUE test set. For both models, we use English Wikipedia to pre-train for three epochs from scratch to be consistent with \citet{tinybert}. Note that the intermediate expansion ratio in our search space is applied to the original intermediate hidden size of TinyBERT$_6$ (i.e., 3072). The results are shown in Table \ref{tab: ablation of architecture transfer}. From the results, we can observe that our EfficientBERT with the base model of TinyBERT$_6$ outperforms the original TinyBERT$_6$ on most of the downstream tasks, and has gained an improvement of 1.3 on the average GLUE score, showing the strong transferability of our EfficientBERT.

\begin{figure}
\includegraphics[trim=80 150 10 230, width=1.\columnwidth, clip]{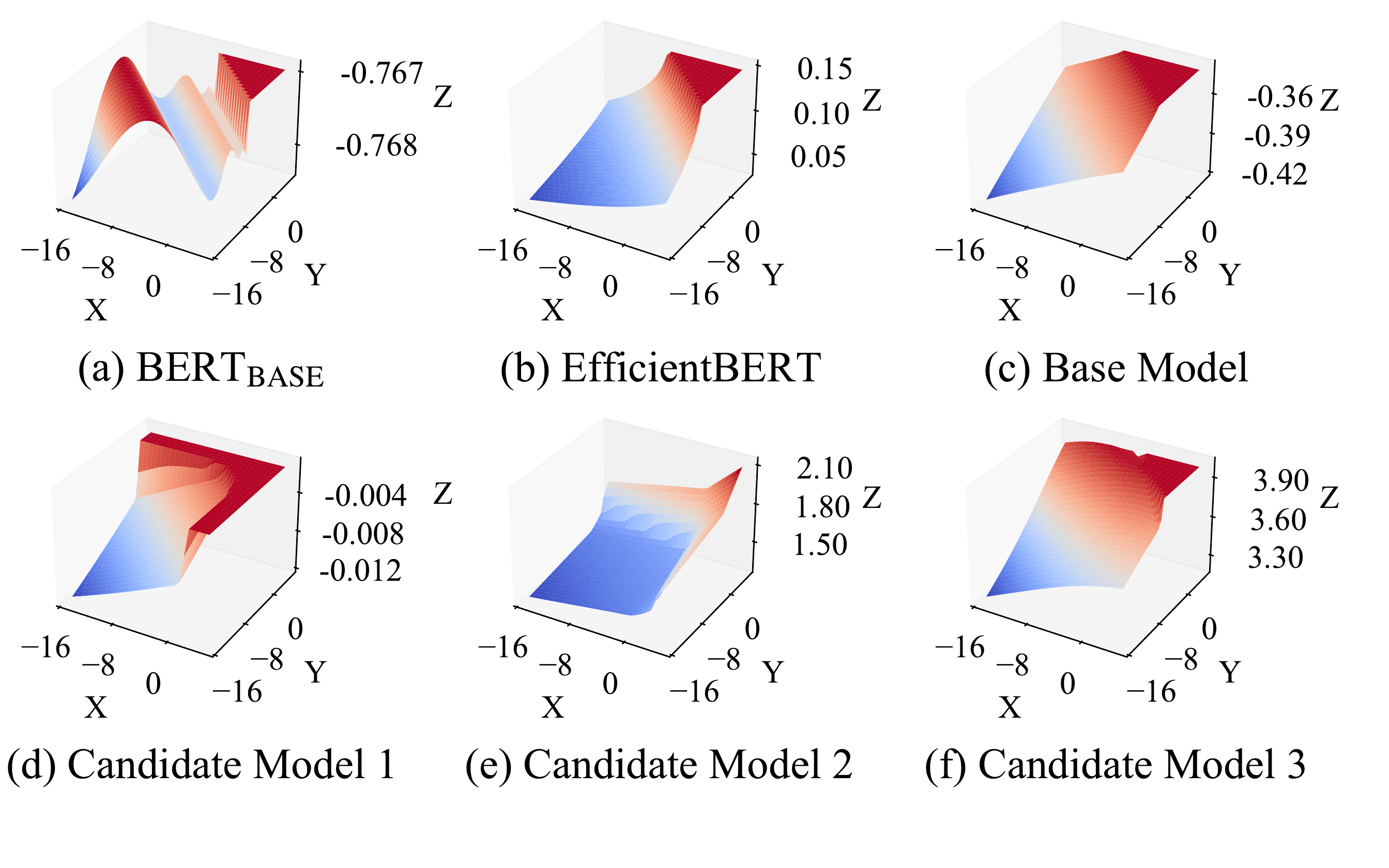}
\caption{Visualization towards the FFN nonlinearity of (a) BERT$\rm _{BASE}$, (b) our EfficientBERT, (c) our base model, and (d)-(f) randomly selected candidate models with worse performances in the first search stage.}\label{fig: ffn nonlinearity}
\end{figure}

\paragraph{Visualization of FFN Nonlinearity.} In Figure \ref{fig: ffn nonlinearity}, we typically visualize the FFN nonlinearity of BERT$\rm _{BASE}$, our EfficientBERT, our base model, and three randomly selected candidate models with worse performances in the first search stage. The input embedding of each model has two dimensions serving as axes X and Y, whose values are uniformly selected from -15$\sim$5 to approximate the distribution of the embedding in BERT$\rm _{BASE}$. The average output of the last Transformer layer is regarded as the value of axis Z. Besides, we remove the MHA, replace the layer normalization with the simple average operation, and set the weights and bias in each linear operation to 1 and 0, respectively, in order to alleviate their impacts. From the results, we can observe that the curves of (a)-(c) are more fluent and have less sudden increase regions than (d)-(f); and from (a) to (c), the curve complexity gradually decreases. It reflects that BERT$\rm _{BASE}$ (our teacher model) has the best FFN nonlinearity, and our EfficientBERT has better nonlinearity than the base model and the randomly selected candidate models. This verifies the superiority of our method in gaining better nonlinear mapping ability. More visualization of the nonlinearity can be seen in Figure \ref{fig: appendix_last_attn_viz}-\ref{fig: appendix_last_ffn_viz} of the Appendix \ref{sec: appendix_2}.

\section{Conclusion}

In this paper, we focus on the compression and improvement of FFN and design a profound search space over the nonlinearity of MLP in FFN, aiming at searching for better MLP architectures to improve the model performance. Due to the enormous search space, we conduct NAS in a progressive manner and employ a novel warm-up KD strategy at each search stage to accelerate searching and enhance model transferability. Extensive experiments show that our searched architecture EfficientBERT is 6.9$\times$ smaller and 4.4$\times$ faster than BERT$\rm _{BASE}$, and has competitive performance and strong generalization ability. In the future, we will leverage NAS to discover more dynamic PLMs \emph{w.r.t} different hardwares and downstream tasks.

\section*{Acknowledgements}

This work was supported in part by National Key R\&D Program of China under Grant No. 2020AAA0109700, National Natural Science Foundation of China (NSFC) under Grant No.U19A2073 and No.61976233, Guangdong Province Basic and Applied Basic Research (Regional Joint Fund-Key) Grant No.2019B1515120039, Guangdong Outstanding Youth Fund (Grant No. 2021B1515020061), Shenzhen Fundamental Research Program (Project No. RCYX20200714114642083, No. JCYJ20190807154211365).

\bibliography{emnlp2021}
\bibliographystyle{acl_natbib}

\appendix

{\large{\noindent \textbf{Appendix}}}

\section{Visualization of Searched Models}
\label{sec: appendix_1}
\begin{figure*}
\centering
\includegraphics[trim=0 60 0 120, width=0.88\textwidth, clip]{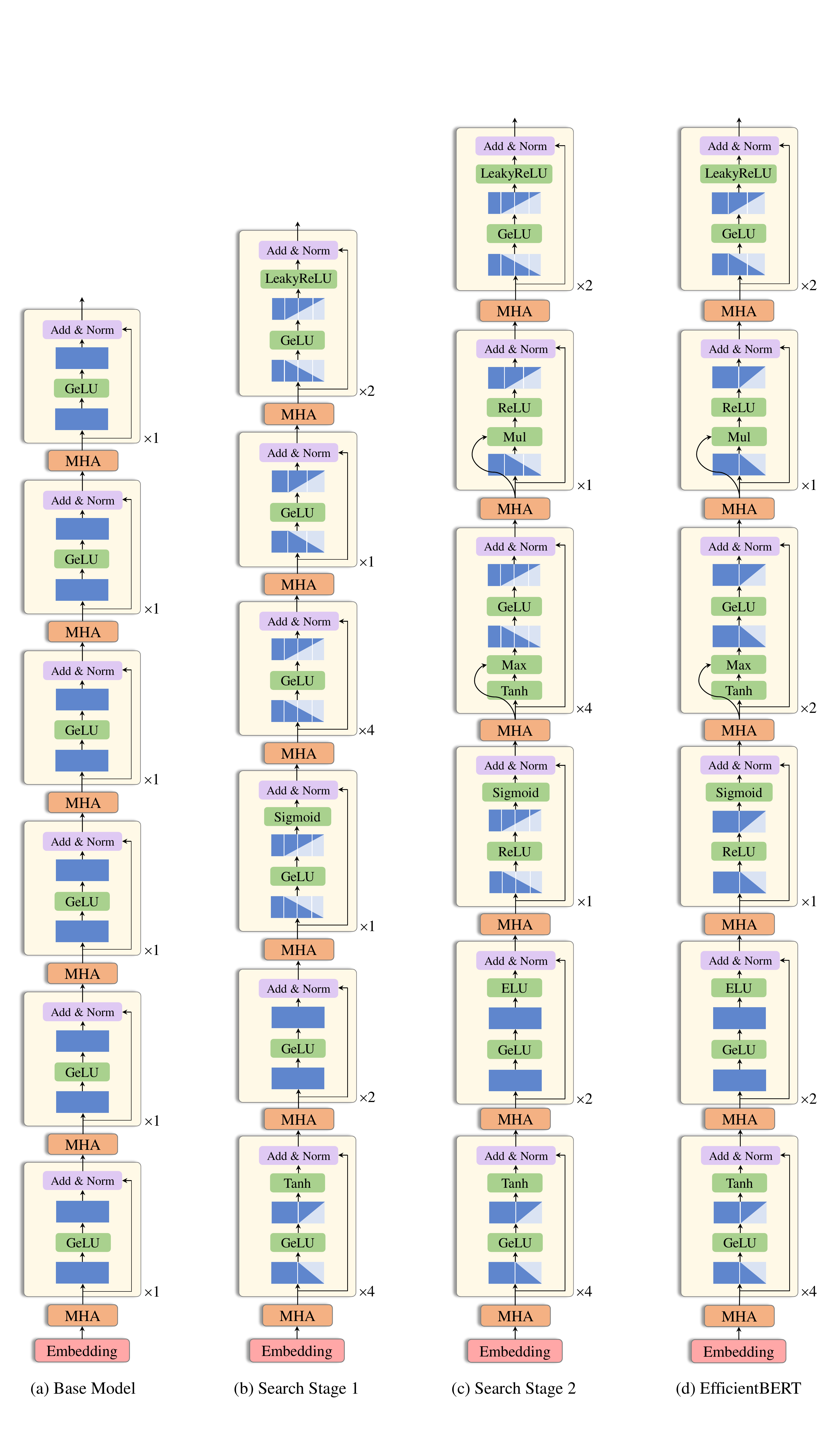}
\caption{Architectures of (a) our base model, (b)-(c) the searched models of the first two search stages, and (d) our EfficientBERT.}
\label{fig: appendix_searched_models}
\end{figure*}

We visualize the architectures of our base model, the searched models of the first two search stages, and our EfficientBERT in Figure \ref{fig: appendix_searched_models} from (a) to (d). From the architecture, we can observe that our EfficientBERT is more efficient since most of the searched intermediate expansion ratios are 1/2 while most of the searched stack numbers are less than 2. Besides, in our EfficientBERT, lower layers tend to have more FFN stack number or intermediate expansion ratio (e.g., layer 1, 2) so as to enrich the semantic representation to the maximum extent for processing by higher layers. In comparison, higher layers tend to learn more complex mathematical formulas (e.g., layers 4, 5) to enhance the nonlinearity of lower enriched representations. This could bring many inspirations for efficient and effective backbone design.

\section{Visualization of Model Nonlinearity}
\label{sec: appendix_2}

\begin{figure*}
\centering
\includegraphics[trim=120 30 100 50, width=\textwidth, clip]{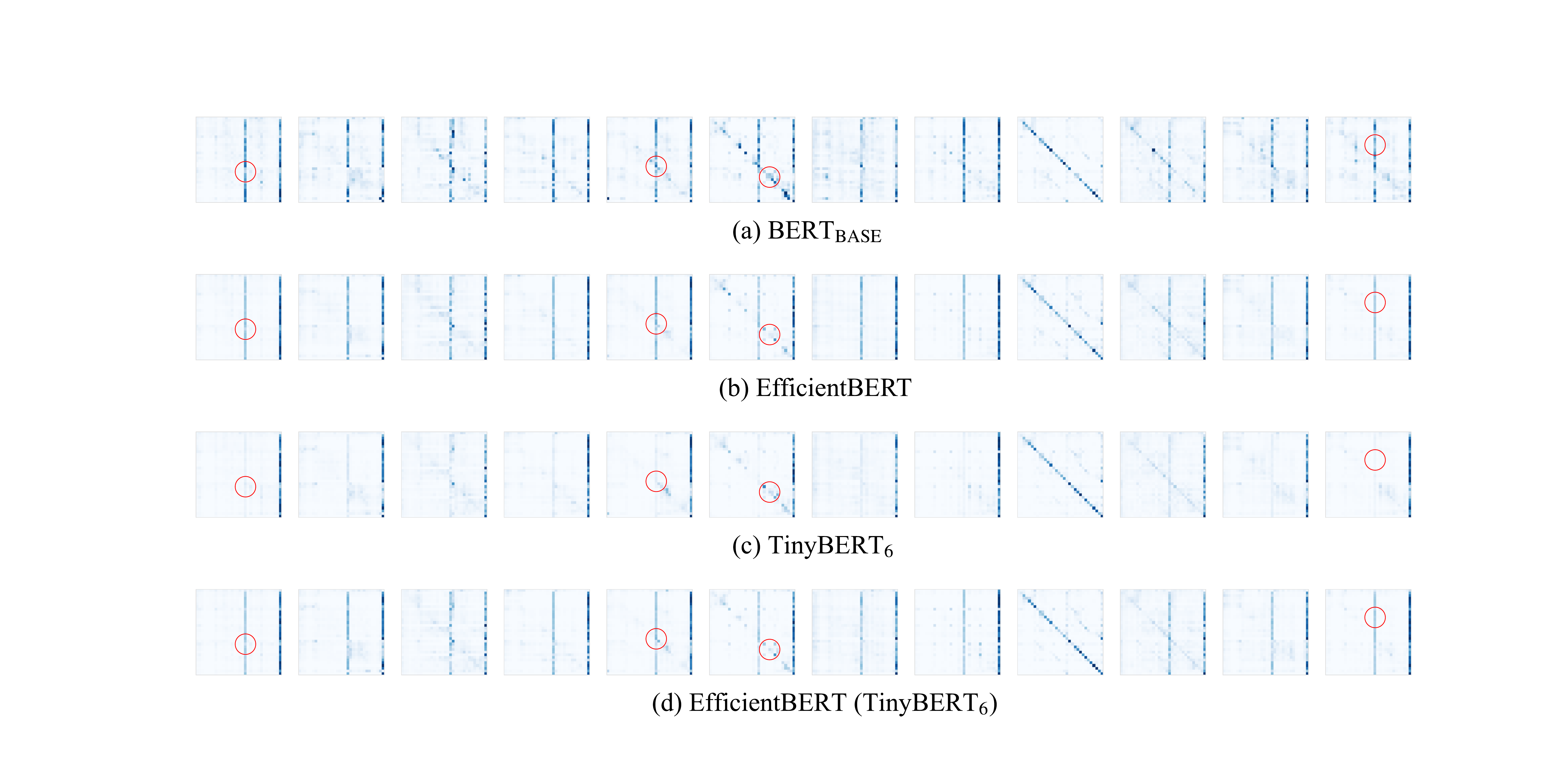}
\caption{Visualization for the attention distributions of (a) BERT$\rm _{BASE}$, (b) our EfficientBERT, (c) TinyBERT$_6$, and (d) our EfficientBERT (TinyBERT$_6$) in the last Transformer layer.}
\label{fig: appendix_last_attn_viz}
\end{figure*}

To further show the superior nonlinearity of our searched models, we visualize the attention maps in twelve attention heads for BERT$\rm _{BASE}$, our EfficientBERT, TinyBERT$_6$, and our EfficientBERT (TinyBERT$_6$) in Figure \ref{fig: appendix_last_attn_viz}, respectively. As can be seen, the feature maps of our EfficientBERT are close to those of BERT$\rm _{BASE}$. This verifies the nonlinear mapping ability of our EfficientBERT in fitting the teacher model. Moreover, the attention distributions of our EfficientBERT (TinyBERT$_6$) are closer to BERT$\rm _{BASE}$ than TinyBERT$_6$ in most of the attention heads. This proves the excellent nonlinear representation ability of our EfficientBERT (TinyBERT$_6$) again.

\begin{figure*}
\centering
\includegraphics[trim=0 0 0 0, width=\textwidth, clip]{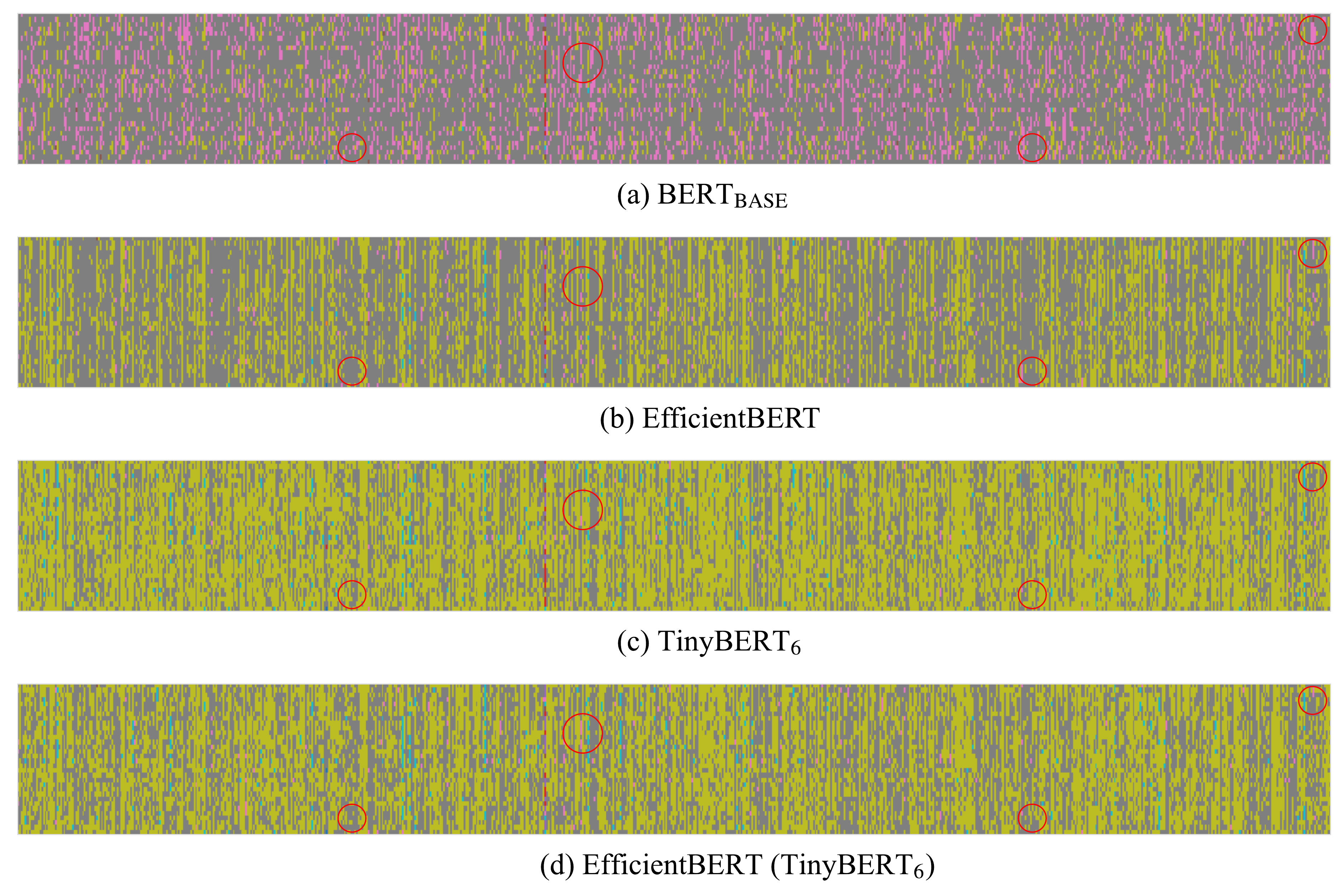}
\caption{Visualization for the FFN output distributions of (a) BERT$\rm _{BASE}$, (b) our EfficientBERT, (c) TinyBERT$_6$, and (d) our EfficientBERT (TinyBERT$_6$) in the last Transformer layer.}
\label{fig: appendix_last_ffn_viz}
\end{figure*}

Then, we visualize the feature maps of FFN outputs for the above four models, as shown in Figure \ref{fig: appendix_last_ffn_viz}. The observations in Figure \ref{fig: appendix_last_ffn_viz} are similar to that of Figure \ref{fig: appendix_last_attn_viz}, once again demonstrating the superior nonlinear representation ability of our EfficientBERT and EfficientBERT (TinyBERT$_6$).\\\\

\end{document}